\pdfoutput=1
\documentclass[sigconf]{acmart}

\usepackage{multirow}

\AtBeginDocument{%
  }

\renewcommand\footnotetextcopyrightpermission[1]{}

\begin{document}

\title{Step Back to Move Forward: Reflection-Aware Preference Optimization for Visual Generation}

\author{Junlong Wu}
\authornote{Both authors contributed equally to this research.}
\email{wu-jl24@mails.tsinghua.edu.cn}
\orcid{0009-0000-9744-1045}
\affiliation{%
  \institution{Tsinghua University}
  \city{Beijing}
  \country{China}
}

\author{Jiuzhou Lin}
\authornotemark[1]
\email{lin-jz24@mails.tsinghua.edu.cn}
\orcid{0009-0005-7730-0479}
\affiliation{%
  \institution{Tsinghua University}
  \city{Beijing}
  \country{China}
}

\author{Jia Sun}
\correspondingauthor
\authornote{Corresponding author.}
\email{sunjia05@kuaishou.com}
\orcid{0009-0009-7688-5726}
\affiliation{%
  \institution{Kuaishou Technology}
  \city{Beijing}
  \country{China}
}

\author{Boheng Zhang}
\email{yangxiao16@kuaishou.com}
\orcid{0000-0003-1185-3239}
\affiliation{%
  \institution{Kuaishou Technology}
  \city{Beijing}
  \country{China}
}

\author{Huaiqing Wang}
\email{wanghuaiqing@kuaishou.com}
\orcid{0009-0007-0429-0948}
\affiliation{%
  \institution{Kuaishou Technology}
  \city{Beijing}
  \country{China}
}

\author{Dewen Fan}
\email{fandewen@kuaishou.com}
\orcid{0000-0002-6062-4208}
\affiliation{%
  \institution{Kuaishou Technology}
  \city{Beijing}
  \country{China}
}

\author{Houde Liu}
\correspondingauthor
\authornotemark[2]
\email{liu.hd@sz.tsinghua.edu.cn}
\orcid{0000-0002-7314-3366}
\affiliation{%
  \institution{Tsinghua University}
  \city{Beijing}
  \country{China}
}

\author{Qianqian Gan}
\email{ganqianqian@kuaishou.com}
\orcid{0009-0003-1812-8993}
\affiliation{%
  \institution{Kuaishou Technology}
  \city{Beijing}
  \country{China}
}

\author{Fan Yang}
\email{yangfan@kuaishou.com}
\orcid{0009-0005-4570-5885}
\affiliation{%
  \institution{Kuaishou Technology}
  \city{Beijing}
  \country{China}
}

\author{Tingting Gao}
\email{gtt0511@163.com}
\orcid{0009-0003-0310-7751}
\affiliation{%
  \institution{Kuaishou Technology}
  \city{Beijing}
  \country{China}
}

\renewcommand{\shortauthors}{Junlong Wu, Jiuzhou Lin et al.}

\begin{abstract}
  Diffusion models have become the mainstream paradigm for modern visual generation and have substantially advanced multimedia content synthesis, especially in text-to-image and text-to-video tasks. To further align such generative models with human preferences, reinforcement learning (RL) has recently shown strong potential as a post-training strategy. Nevertheless, existing policy gradient-based methods often explore inefficiently, making them vulnerable to local optima that may degrade semantic faithfulness and visual realism. To address these challenges, we present Reflection-Aware GRPO (RA-GRPO), a new RL-based preference alignment framework for diffusion generative models. The core idea is to improve `forward' generation by incorporating `backward' reflection during optimization. We first introduce Diffusion Reflection, which rectifies intermediate sampling trajectories by inverting the diffusion process with a weak estimator, guiding latent states toward higher-probability regions of the true data manifold. Furthermore, we introduce Counterfactual Path Synthesis to implicitly distill these rectified trajectories into the policy, enabling the model to internalize the benefits of search-based exploration without incurring inference-time overhead. Extensive experiments on T2I and T2V models demonstrate that RA-GRPO significantly outperforms existing methods, particularly in mitigating reward hacking and improving generalization. The method remains architecture-agnostic and integrates seamlessly with standard pipelines, suggesting a promising direction for stable preference alignment.
\end{abstract}




\maketitle
\thispagestyle{empty}
\pagestyle{plain}

\section{Introduction}
Recent advances in diffusion models \cite{ho2020denoising, song2020denoising, song2020score, lipman2022flow, podell2023sdxl, blackforestlabs2024flux, wan2025wan} have significantly advanced visual generation and enabled high fidelity image synthesis from textual descriptions. As diffusion models become a mainstream paradigm for multimedia content creation, aligning their outputs with complex human preferences, such as aesthetic quality, semantic consistency, and structural faithfulness, has become increasingly important. In this context, reinforcement learning has emerged as a promising post training strategy for preference alignment. Standard reinforcement learning objectives \cite{schulman2017proximal, ouyang2022training} formulate alignment as the maximization of expected reward, where policy updates are driven by gradient estimates computed from trajectories sampled from the model's current policy.

However, this dependency exposes an important limitation. The optimization process of RL is largely restricted to the current generative manifold of the model \cite{casper2023open, rafailov2023direct}. In practice, the model mainly improves by reweighting already accessible generation paths, rather than exploring potentially better inference trajectories or knowledge regions that remain outside its current capability. This limitation is especially problematic for Group Relative Policy Optimization (GRPO) \cite{shao2024deepseekmath}, which relies on relative advantage estimation, particularly during the early cold start stage. At this stage, the model is still poorly aligned, and the generated samples for a given prompt are often uniformly low in quality.  Under such circumstances, DanceGRPO \cite{xue2025dancegrpo} uses the group average as the baseline and treats samples that are only relatively better within the group as positive signals. For visual generation tasks with sparse rewards, this strategy is often ineffective, since the selected samples may still be far from truly desirable outputs in terms of semantic fidelity and visual quality. Without genuinely high reward samples to provide reliable guidance, the resulting gradient estimates become noisy, exploration remains inefficient, and the optimization process is prone to poor local optima.

To address these limitations, we incorporate a novel exploration mechanism, termed Diffusion Reflection, into the trajectory sampling process. Rather than restricting exploration to the forward trajectories induced by the current policy, our method introduces an inverse diffusion process to refine sampled paths during training. By alternating denoising and inversion operations, Diffusion Reflection redirects latent variables away from suboptimal trajectories and toward regions that are better aligned with the true data distribution and associated with higher rewards. In this way, the proposed mechanism improves sample quality, alleviates the exploration bias of the current policy, and expands the range of trajectories available for effective preference optimization.

Although Diffusion Reflection produces improved trajectories, directly exploiting these trajectories for policy optimization is nontrivial. The main challenge is not sample validity, but the mismatch between reflection refined trajectories and the training signals used in standard policy gradient methods. In conventional optimization, learning is driven by the log probabilities of transitions sampled from the current policy. As a result, improvements introduced by reflection are difficult to absorb effectively, since the refined states are not explicitly represented as natural transition outcomes under the original sampling path. Consequently, the policy may benefit from better samples at the reward level, yet still fail to learn the transition behavior that reproduces these improvements during generation. To address this issue, we propose Counterfactual Path Synthesis, a mechanism that converts reflection refined trajectories into consistent supervision for policy learning. Instead of treating the refined trajectory as an isolated search result, we construct a counterfactual transition that associates the refined state with the original generation context and forms a coherent training path for optimization. This design enables the policy to internalize the useful structure revealed by reflection and gradually reproduce similar improvements through standard generation. In this way, the benefit of reflection based search is distilled into the model during training, without introducing additional cost during inference.

In summary, our contributions are as follows:
\begin{itemize}
\setlength{\itemsep}{0pt}
\item We introduce Diffusion Reflection, a new exploration mechanism for RL based optimization of diffusion models. By exploiting the invertibility of diffusion dynamics, it guides trajectory sampling toward more promising high reward regions.
\item We propose Counterfactual Path Synthesis, a training strategy that bridges the gap between reflection refined trajectories and standard policy optimization. This design enables the model to absorb the benefits of reflection based search through implicit distillation.
\item We conduct extensive experiments to validate the proposed method. The results show that Reflection Aware GRPO consistently outperforms strong baselines in both alignment efficiency and generation quality.
\end{itemize}

\section{Related Work}
\subsection{RL-based Alignment for Generative Models}
Aligning generative models with human preferences has evolved from supervised fine-tuning to sophisticated Reinforcement Learning (RL) paradigms, following their widespread and successful adoption in Large Language Models (LLMs) \cite{luo2025o1, lee2023rlaif, ouyang2022training, bai2022training, touvron2023llama}. Notable works \cite{black2023training, xu2023imagereward, fan2023dpok} treat the denoising process as a multi-step decision-making problem, optimizing the score function to maximize aesthetic scores or groundability. Methods \cite{wallace2024diffusion, sun2025generalizing} introduced an offline approach adapted from DPO \cite{fan2023reinforcement}, allowing models to learn directly from paired preference data. Recent advancements \cite{xue2025dancegrpo, liu2025flow} have successfully transplanted GRPO \cite{shao2024deepseekmath} to continuous-time generative models. These methods typically transform the deterministic Ordinary Differential Equation (ODE) sampling into a Stochastic Differential Equation (SDE) formulation at each timestep. To further refine the efficacy of GRPO, recent variants \cite{li2025mixgrpo, he2025tempflow, luo2025sample, wang2025pref, guo2025g, li2025branchgrpo} have focused on structural optimizations regarding sampling efficiency and update granularity. By employing hybrid ODE-SDE sampling strategies, decomposing long-horizon trajectories into manageable segments, or refining the scope of supervision, these methods effectively mitigate the computational overhead and temporal credit assignment ambiguity inherent in full-step SDE sampling.  Beyond visual generation, diffusion models and reinforcement learning have been widely applied to
  diverse perception and control problems~\cite{li2025diffpcn, yan2025symmcompletion, li2026detailanywhere, gong2026sculpting, wu2025arc, li2025safesim, lin2025keypoint}, underscoring the generality of reward-driven
  optimization as a paradigm.

\begin{figure*}[t]
    \centering
    \includegraphics[width=\textwidth]{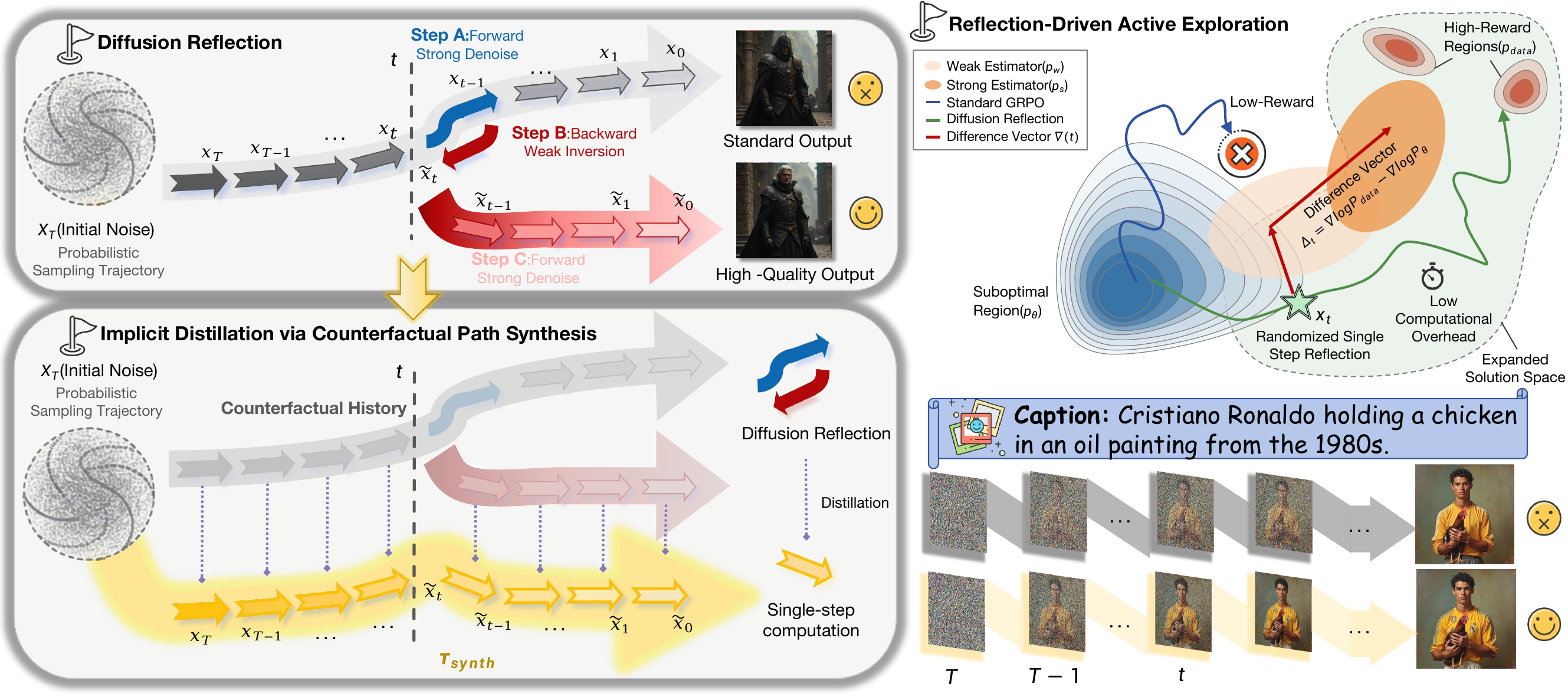}
    \caption{Visualization of the proposed RA-GRPO framework. The figure illustrates the key components of the RA-GRPO pipeline, including Diffusion Reflection, Implicit Distillation via Counterfactual Path Synthesis, and Active Exploration. The process involves a probabilistic sampling trajectory, with steps of forward denoising, backward reflection, and counterfactual path synthesis. The system aims to generate high-quality outputs while exploring expanded solution spaces with low computational overhead. Example output: A synthetic oil painting of Cristiano Ronaldo holding a chicken from the 1980s.}
    \label{fig:arch}
\vspace{-0.3cm}
\end{figure*}

\subsection{Self-Reflection in Generative Models}
Drawing inspiration from the success of reflective mechanisms in Large Language Models, where models iteratively critique and refine their own outputs to enhance reasoning capabilities \cite{madaan2023self, shinn2023reflexion, pan2023automatically}, recent research has adapted similar self-corrective paradigms to diffusion-based visual generation. Existing approaches largely branch into two paradigms: (i) Inference-Time Refinement, where methods \cite{li2025reflect, bai2024zigzag, bai2025weak} enhance generation fidelity by iteratively refining latent representations or utilizing non-monotonic forward-backward sampling strategies to dynamically adjust the generative path; (ii) Optimization via Feedback \cite{lyu2025realrag, zhuo2025reflection}, which uses reflective signals as supervision to distill teacher-level capabilities into weaker models or to fine-tune responses to complex prompts. Despite their effectiveness, these approaches typically treat reflection as a temporary inference-time procedure and lack a mechanism to convert this additional computation into a persistent training-time signal that shapes the model’s generative behavior.

\section{Method}
In this section, we present our proposed framework, \textbf{R}eflection-\textbf{A}ware \textbf{G}roup \textbf{R}elative \textbf{P}olicy \textbf{O}ptimization (\textbf{RA-GRPO}). We first formulate reinforcement learning for diffusion models. Next, we introduce Diffusion Reflection, an exploration mechanism that improves trajectory sampling by refining latent trajectories toward more promising regions. We then present Counterfactual Path Synthesis, which converts reflection-refined trajectories into effective supervision for policy optimization. Finally, we discuss how the proposed framework helps reduce reward hacking during training. An overview of RA-GRPO is illustrated in Figure~\ref{fig:arch}.

\subsection{Preliminaries: RL for Diffusion Models}

Flow Matching~\cite{lipman2022flow} learns a continuous-time vector field that transports samples between a simple prior distribution and the data distribution \(p_{\mathrm{data}}\). Let \(x_0 \sim p_{\mathrm{data}}\) denote a real data sample and \(x_1 \sim \mathcal{N}(0, I)\) denote a noise sample. Following the conditional optimal transport formulation, we define a linear probability path as:
\begin{equation}
    x_t = (1 - t)x_0 + t x_1, \quad t \in [0, 1].
\end{equation}
The corresponding conditional target velocity along this path is
\begin{equation}
    u_t(x_t \mid x_0, x_1) = \frac{d x_t}{dt} = x_1 - x_0,
\end{equation}
and the model is trained to learn a time-dependent velocity field \(v_\theta(x_t, t)\) that matches this transport behavior.

At inference time, Flow Matching typically performs deterministic sampling by solving the ordinary differential equation
    $\frac{d x_t}{dt} = v_\theta(x_t, t)$.
While this deterministic formulation is effective for generation, it provides limited exploration for reinforcement learning, since the transition from a given state is fully determined by the current vector field. In contrast, GRPO requires multiple diverse rollouts under the same conditioning input in order to estimate group-wise relative advantages. To introduce stochasticity into the sampling process, we follow recent works~\cite{xue2025dancegrpo, liu2025flow} and adopt an equivalent stochastic differential equation formulation:
\begin{equation}
\label{eq:sde_sampling}
    d x_t =
    \left[
    v_\theta(x_t, t)
    + \frac{\sigma_t^2}{2t}\Big(x_t + (1-t)v_\theta(x_t, t)\Big)
    \right] dt
    + \sigma_t d w_t,
\end{equation}
where \(w_t\) denotes a Wiener process and \(\sigma_t\) is a predefined noise schedule controlling the level of stochasticity. In practice, we solve the SDE over \(t \in [\epsilon, 1]\) with a small \(\epsilon > 0\) to avoid the singularity at \(t=0\). Under this construction, the correction term compensates for the injected noise and preserves the marginal distributions of the underlying deterministic flow.

Given a prompt \(c\), the model generates a group of \(G\) trajectories \(\{\tau^{(i)}\}_{i=1}^G\) using the stochastic sampler described above, where \(\tau^{(i)} = \{x_t^{(i)}\}_{t=0}^T\) and \(x_0^{(i)}\) denotes the final generated sample. We compute a trajectory-level reward \(r(x_0^{(i)}, c)\) for each sample and normalize it within the group to obtain the relative advantage:
\begin{equation}
    A_i =
    \frac{
        r(x_0^{(i)}, c)
        -
        \frac{1}{G}\sum_{j=1}^G r(x_0^{(j)}, c)
    }{
        \mathrm{std}\big(\{r(x_0^{(j)}, c)\}_{j=1}^G\big) + \delta
    },
\end{equation}
where \(\delta\) is a small constant for numerical stability. Since the reward is defined on the final generated sample, the resulting advantage \(A_i\) is shared across all timesteps of the \(i\)-th trajectory.

The policy parameters \(\theta\) are then optimized with the GRPO objective:
\begin{equation}
\begin{aligned}
\mathcal{J}_{\mathrm{GRPO}}(\theta)
=
\mathbb{E}_{c,\{\tau^{(i)}\}_{i=1}^G \sim \pi_{\mathrm{old}}}
\Bigg[
\frac{1}{G}\sum_{i=1}^G \frac{1}{T}\sum_{t=1}^T
\Big(
\min \big(
\rho_t^{(i)} A_i,\,
\\
\mathrm{clip}(\rho_t^{(i)}, 1-\varepsilon, 1+\varepsilon) A_i
\big)
- \beta\,
\mathbb{D}_{\mathrm{KL}}
\big(
\pi_\theta
\,\|\, 
\pi_{\mathrm{ref}}
\big)
\Big)
\Bigg],
\end{aligned}
\end{equation}
where
$
    \rho_t^{(i)}
    =
    \frac{
        \pi_\theta(x_{t-1}^{(i)} \mid x_t^{(i)}, c, t)
    }{
        \pi_{\mathrm{old}}(x_{t-1}^{(i)} \mid x_t^{(i)}, c, t)
    }
$. Here, \(\varepsilon\) is the PPO clipping parameter, and \(\beta\) controls the strength of KL regularization with respect to the reference policy \(\pi_{\mathrm{ref}}\), which helps stabilize policy updates and prevent excessive drift from the pretrained model.

\subsection{Active Exploration via Diffusion Reflection}
\label{section3.2}
A fundamental limitation of GRPO is that exploration is restricted to the effective support of the current policy \(\pi_\theta\). In the cold-start stage, \(\pi_\theta\) typically assigns most probability mass to suboptimal regions, so high-reward modes under the target distribution remain difficult to access through standard on-policy sampling. As a result, policy improvement is often driven by selecting relatively better samples from an overall low-quality group, rather than by discovering genuinely superior trajectories.

To mitigate this issue, we introduce \emph{Randomized Single-Step Reflection} as an active exploration mechanism during sampling, motivated by diffusion reflection ~\cite{bai2024zigzag, bai2025weak}. Prior work has demonstrated that diffusion reflection serves as an effective training-free technique that improves sampling quality, partly by reducing the gap between the model-induced estimator and the true data distribution. In our setting, the weak--strong construction is instantiated directly through \emph{text-guidance level}: for the same latent state \(x_t\), timestep \(t\), and condition \(c\), we evaluate the same velocity model \(v_\theta\) under two guidance scales \(w_{\mathrm{w}} < w_{\mathrm{s}}\), and define
$
    v_\theta^{\mathrm{w}}(x_t, t, c) := v_\theta(x_t, t, c; w_{\mathrm{w}}) 
$,
$
    v_\theta^{\mathrm{s}}(x_t, t, c) := v_\theta(x_t, t, c; w_{\mathrm{s}}).
$
Here, the weak and strong estimators correspond to the implicit conditional distributions induced by weaker and stronger text guidance, respectively. Under the assumption that stronger guidance produces samples that are better aligned with the prompt and thus closer to the target conditional data distribution $p_{\mathrm{data}}(\cdot \mid c)$, the discrepancy between the two estimators,
$
    \Delta_{\mathrm{ws}}(x_t, t, c)
    :=
    v_\theta^{\mathrm{s}}(x_t, t, c) - v_\theta^{\mathrm{w}}(x_t, t, c),
$
can be interpreted as a first-order correction direction. Following the weak-to-strong perspective can be interpreted as an empirical approximation to the missing score correction up to approximation error:
\begin{equation}
    \Delta_{\mathrm{ws}}(x_t, t, c)
    \;\propto\;
    \nabla_{x_t}\log p_{\mathrm{data}}(x_t \mid c)
    -
    \nabla_{x_t}\log p_\theta(x_t \mid c).
\end{equation}
 Injecting this correction during sampling therefore perturbs trajectories away from regions overly favored by the current policy and steers them toward higher-reward regions under \(p_{\mathrm{data}}\).

A direct application of such reflection at every denoising step is computationally expensive. We therefore adopt a lightweight strategy termed \emph{Randomized Single-Step Reflection}. For a reflected trajectory, we first sample a timestep \(t_r \sim \mathcal{U}[0.2T, T]\), and perform the reflection operation only once at \(t_r\). Concretely, given the current latent state \(x_{t_r}\), we execute a \emph{forward--backward--forward} update:
\begin{itemize}
\item Step A: We first project the latent state $x_{t_r}$ forward using the strong velocity field $v^{\mathrm{s}}_\theta$. This estimates a cleaner state $x_{t_r-1}$:
$
    x_{t_r-1}
    =
    x_{t_r}
    + v^{\mathrm{s}}_\theta(x_{t_r}, t_r, c)\,\Delta t
$.

\item Step B: We then invert the state back to the current timestep using a weak velocity field $v^{\mathrm{w}}_\theta$. This inversion pulls the state back along an alternative direction, providing a corrective adjustment:
$
    \tilde{x}_{t_r}
    =
    x_{t_r-1}
    - v^{\mathrm{w}}_\theta(x_{t_r-1}, t_r-1, c)\,\Delta t.
$

\item Step C: Finally, we resume the standard sampling process from this rectified state $\tilde{x}_{t_r}$, applying the strong velocity field again to proceed to $t_r-1$:$\tilde{x}_{t_r-1}
=
\tilde{x}_{t_r}
+ v^{\mathrm{s}}_\theta(\tilde{x}_{t_r}, t_r, c)\,\Delta t.$
\end{itemize}

Importantly, Diffusion Reflection is not applied to every trajectory within a group. Given a group of \(G\) sampled trajectories, we randomly select only a fraction \(r \in (0,1]\) to perform the reflection operation. This partial-reflection design naturally induces \emph{active exploration}: rather than passively ranking samples confined to the current policy support, the reflection injects a correction into selected trajectories, perturbing them toward regions of higher probability density under \(p_{\mathrm{data}}\) and thereby encouraging the discovery of higher-quality modes that are otherwise difficult to reach. As illustrated in Figure~\ref{fig:arch}, the ``Diffusion Reflection'' and ``Reflection-Driven Active Exploration'' components depict the latent-space reflection process and show how the corrected trajectories are redirected toward more desirable regions beyond the original policy's effective support. At the same time, applying reflection to only a subset of trajectories introduces only a small additional computational overhead, whose impact will be analyzed in the experimental section. Overall, Randomized Single-Step Reflection provides an efficient and targeted mechanism for active exploration, enabling the policy to escape poor local regions and discover better trajectories while remaining fully compatible with group-based policy optimization.

\subsection{Implicit Distillation via Counterfactual Path Synthesis}
While reflected samples $\tilde{x}_0$ serve as high-quality outputs, they are generated via a complex, non-monotonic inference process. As detailed in Section \ref{section3.2}, a reflection trajectory $ \tau_{ref} $ involves a Single-Step Reflection at a randomly selected timestep $t_r$: 
\begin{equation}
    \tau_{ref} = (x_T, \dots, x_{t_r}, x_{{t_r}-1} \xrightarrow{\text{invert}} \tilde{x}_{t_r}, \tilde{x}_{{t_r}-1}, \dots, \tilde{x}_0)
\end{equation}
The presence of this reflection step makes the reflected trajectory non-standard and therefore not directly compatible with the monotonic one-pass ODE trajectory used during efficient inference. To bridge this gap, we propose Counterfactual Path Synthesis. The key intuition is that, if the policy were sufficiently optimal, it should be able to reach the same high-reward sample from the state $x_{t_r}$ through a standard forward denoising path, without requiring the additional corrective reflection step.

\textbf{Construction of the Synthesized Trajectory} We construct a hybrid training trajectory $\tau_{synth}$ by stitching the trajectory before the reflection step with the trajectory after the reflection step. Formally: $\tau_{synth} = \left( x_T, \dots, x_{{t_r}+1}, \tilde{x}_{t_r}, \dots, \tilde{x}_0 \right)$, where the segment $(x_T, \dots, x_{t_r+1})$ preserves the original stochastic path sampled from Gaussian noise, and the segment $(\tilde{x}_{t_r}, \dots, \tilde{x}_0)$ represents the high-fidelity path derived from the reflection operation. This construction creates a counterfactual history: it represents the trajectory the model should have taken at the critical branching point $t_r$ to reach the optimal outcome.

\textbf{Optimization as Implicit Distillation.}
We retain the GRPO backbone and introduce a targeted intervention only at the
reflection timestep~$t_r$. At this step the model is encouraged to steer toward
the synthesized state~$\tilde{x}_{t_r}$, weighted by the advantage of the
resulting outcome~$\tilde{x}_0$. Because $\tilde{x}_{t_r}$ is produced by an
external weak-to-strong guidance procedure rather than the current policy, the
objective takes the form of advantage-weighted maximum likelihood:
\begin{equation}
\mathcal{J}_{\mathrm{refl}}(\theta)
=
\mathbb{E}_{c,\, x_{t_r+1},\, t_r,\, \tilde{x}_{t_r}}\Big[
A(\tilde{x}_0)\,
\log q_\theta\big(\tilde{x}_{t_r} \mid x_{t_r+1}, c\big)
\Big].
\end{equation}

We instantiate $q_\theta$ via the velocity field $v_\theta$. Under an Euler
discretization of the probability flow ODE, the one-step transition is an
isotropic Gaussian:
$ q_\theta\big(\tilde{x}_{t_r} \mid x_{t_r+1}, c\big) = \mathcal{N}\Big(
\tilde{x}_{t_r}; x_{t_r+1} + v_\theta(x_{t_r+1}, c, t_{r+1})\,$ 
$\Delta t,
\sigma^2 I \Big),$
where $\Delta t$ is the step size from $t_{r+1}$ to $t_r$. Differentiating
$\mathcal{J}_{\mathrm{refl}}$ under this parameterization shows that the
gradient is proportional to the negative gradient of the following squared-error
loss:
\begin{equation}\label{eq:refl_loss}
\mathcal{L}_{\mathrm{refl}}(\theta)
=
\mathbb{E}\left[
A(\tilde{x}_0)\cdot
\frac{1}{2}
\left\|
v_\theta(x_{t_r+1}, c, t_{r+1})
-
\frac{\tilde{x}_{t_r} - x_{t_r+1}}{\Delta t}
\right\|_2^2
\right],
\end{equation}
where the constants $\sigma^2$ and $(\Delta t)^2$ are absorbed into the learning
rate. In practice we replace $A(\tilde{x}_0)$ with
$\max\!\big(A(\tilde{x}_0),\,0\big)$: negative advantages would otherwise
maximize the squared error, causing gradient instability.

The total training objective combines the standard GRPO loss with the reflection
loss:
\begin{equation}\label{eq:total_loss}
\mathcal{L}(\theta)
=
\mathcal{L}_{\mathrm{GRPO}}(\theta)
\;+\;
\lambda\,\mathcal{L}_{\mathrm{refl}}(\theta).
\end{equation}
Intuitively, $\mathcal{L}_{\mathrm{refl}}$ locally adjusts the model's predicted
velocity at $t_r$ so that the resulting update better matches the
reflection-guided corrective transition. As training proceeds, the external search effort is
progressively absorbed into the policy parameters, enabling the model to
reproduce high-quality trajectories at inference time without explicit reflection.

\subsection{Why Reflection Can Mitigate Reward Hacking}

Reward hacking in generative RL arises when optimizing a misspecified reward drives the policy toward samples that score highly under the target reward model but do not correspond to genuine quality improvements. In standard RL fine-tuning, such shortcut behaviors can become self-reinforcing, since trajectories are repeatedly reinforced as long as they increase the target reward, even if they exploit reward-specific artifacts.

RA-GRPO mitigates this issue through reflection-based local correction. Instead of directly reinforcing an entire high-reward trajectory, we construct a reflected target \(\tilde{x}_{t_r}\) at a selected timestep \(t_r\), derived from a higher-quality synthesized outcome \(\tilde{x}_0\), and train the model to match the local transition \(x_{t_r+1}\rightarrow \tilde{x}_{t_r}\). Because \(\tilde{x}_t\) is obtained by editing the current trajectory rather than replacing it with an arbitrary reward-maximizing sample, the resulting update remains local in trajectory space and is anchored by a better endpoint. This can be viewed as a form of trajectory-level regularization. Conceptually, let \(d_{\mathrm{rew}}\) denote the reward-improving direction favored by standard RL, and let \(d_{\mathrm{ref}}\) denote the reflected corrective direction induced by \(\tilde{x}_{t_r}\). Then the effective update can be interpreted as
\begin{equation}
    d_{\mathrm{update}} \approx d_{\mathrm{rew}} + \lambda d_{\mathrm{ref}},
\end{equation}
where \(\lambda\) is an implicit coefficient representing the effective strength of the reflection-based correction and \(d_{\mathrm{ref}}\) suppresses reward improving directions that require unstable or off-manifold deviations, while preserving directions that admit locally consistent corrections under the denoising dynamics. As a result, RA-GRPO tends to improve reward in a way that better aligns with actual sample quality, which explains its stronger robustness and cross-reward generalization in Table~\ref{tab:reward_model_generalization}.

\begin{table*}[t]
\centering
\small
\setlength{\tabcolsep}{6pt}
\caption{Comparison of different training methods under various reward models evaluated on multiple automatic metrics.}
\begin{tabular}{l l c c c c c }
\toprule
Reward Model & Method
& HPSv2.1 $\uparrow$ & PickScore $\uparrow$ & CLIPScore $\uparrow$ & ImageReward $\uparrow$ & HPSv3 $\uparrow$ \\
\midrule

/ & Flux.1-dev        & 0.305 & 0.230 & 0.388 & 1.127 & 13.683  \\
\midrule
\multirow{3}{*}{HPSv2.1}
& DanceGRPO   & 0.359 & 0.227 & 0.362 & 1.273 & 14.400  \\
& MixGRPO     & 0.370 & 0.227 & 0.363 & 1.377 & 14.717 \\
& \textbf{Ours}
              & \textbf{0.378}
              & \textbf{0.230}
              & \textbf{0.369}
              & \textbf{1.417}
              & \textbf{15.324}  \\
\midrule

\multirow{3}{*}{PickScore}
& DanceGRPO   & 0.323 & 0.239 & 0.376 & 1.162 & 12.626  \\
& MixGRPO     & 0.323 & 0.238 & 0.378 & 1.193 & 12.980  \\
& \textbf{Ours}
              & \textbf{0.332}
              & \textbf{0.241}
              & \textbf{0.389}
              & \textbf{1.322}
              & \textbf{14.251}  \\
\midrule

\multirow{3}{*}{HPSv2.1 \& CLIPScore}
& DanceGRPO   & 0.347 & 0.228 & 0.386 & 1.263 & 14.120  \\
& MixGRPO
              & 0.353
              & 0.229
              & 0.394
              & 1.363
              & 14.272 \\
& \textbf{Ours}
              & \textbf{0.359}
              & \textbf{0.231}
              & \textbf{0.402}
              & \textbf{1.445}
              & \textbf{14.496} \\
\bottomrule
\end{tabular}
\label{tab:reward_model_generalization}
\end{table*}

\begin{figure*}[t!]
    \centering
    \includegraphics[width=0.9\textwidth]{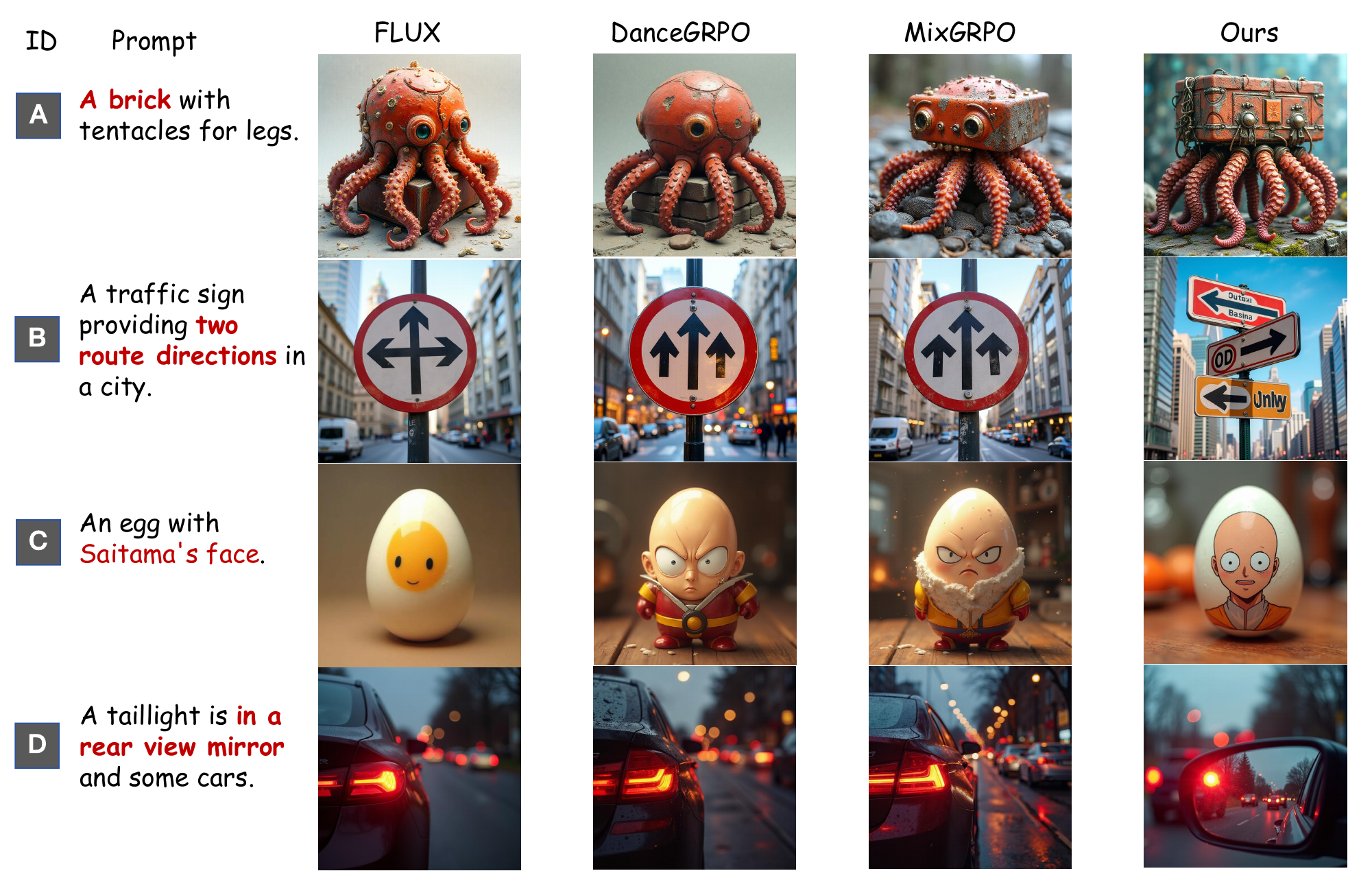}
    \caption{Qualitative comparison of image generation across different methods: Flux.1-dev, DanceGRPO, MixGRPO, and our method. For each prompt (A–D), the generated images highlight the varying degrees of semantic accuracy, creativity, and fidelity, with our method consistently producing better results in terms of visual quality and alignment with the text.}
    \label{fig:qualitative}

\end{figure*}

\section{Experiment}

\subsection{Experiments Setup}

We evaluate RA-GRPO on the HPS v2.1 benchmark \cite{wu2023human}, utilizing prompts designed to rigorously assess human preference alignment. The backbone model is FLUX.1-Dev, a state-of-the-art rectified flow transformer. We compare our method against the base model and two competitive reinforcement learning baselines: DanceGRPO \cite{xue2025dancegrpo}, which introduces stochasticity into deterministic sampling, and MixGRPO \cite{li2025mixgrpo}, employing mixed sampling strategies. To provide a comprehensive evaluation of alignment quality and generalization, we report results across a diverse set of metrics including HPSv2.1 \cite{wu2023human}, PickScore \cite{kirstain2023pickapic}, CLIPScore \cite{hessel2021clipscore}, ImageReward \cite{xu2023imagereward}, and HPSv3 \cite{ma2025hpsv3}. 

\subsection{Implementation Details}

The reflection mechanism is implemented using a guidance-based weak-to-strong pair: since FLUX controls generation through guidance levels rather than standard classifier-free guidance, we set the guidance level of the strong estimator to 3.5 and that of the weak estimator to 1.0. Reflection is applied at a randomized timestep sampled from \(t \sim \mathcal{U}[0.2T, T]\) to construct counterfactual training trajectories. To ensure a fair comparison given the additional computational overhead of our method, our model is trained for 300 steps, while all other methods are trained for 5\% more steps. Training uses AdamW with a learning rate of \(1 \times 10^{-5}\) and weight decay of \(1 \times 10^{-4}\) on 8 \(\times\) NVIDIA H800 GPUs. 

\subsection{Main Results}

\textbf{\textit{Quantitative Evaluation.}} Table~\ref{tab:reward_model_generalization} summarizes alignment performance under different reward optimization objectives, evaluated using multiple automatic metrics. Overall, RA-GRPO demonstrates the strongest balance between optimizing the designated reward and maintaining generalization across heterogeneous evaluators. Under HPSv2.1 optimization, RA-GRPO achieves the highest HPSv2.1 score, while simultaneously obtaining the best results on other evaluation metrics. Under PickScore optimization, RA-GRPO again achieves the strongest overall performance. In the multi-objective setting, RA-GRPO consistently ranks first across all evaluated metrics, further indicating its robustness under joint reward optimization. Taken together, these results suggest a potential reward-hacking phenomenon in generative reinforcement learning: improvements on the optimized reward do not necessarily translate into broad gains under independent evaluation metrics. This limitation is most evident for DanceGRPO, whose improvements on the target reward are not consistently accompanied by stronger performance on other evaluators. By contrast, RA-GRPO exhibits more stable gains across diverse metrics, indicating that it more reliably improves overall sample quality rather than overfitting to reward-specific artifacts.

\textbf{\textit{Qualitative Comparison.}} As illustrated in Figure \ref{fig:qualitative}, our method demonstrates a stronger ability to move beyond the intrinsic generation manifold associated with standard diffusion sampling, thereby yielding improvements in both semantic fidelity and visual diversity. In contrast, baseline methods largely remain concentrated around dominant visual modes acquired during pretraining, which constrains their capacity to faithfully render unconventional concepts. For prompts involving atypical object compositions or attribute bindings (e.g., A and C), baseline methods frequently revert to familiar shapes or stylistic patterns, resulting in incomplete or distorted generations. By comparison, our method produces samples that are more structurally coherent and semantically consistent with the input prompts. Furthermore, for prompts that require precise spatial reasoning (e.g., B and D), our approach exhibits stronger fine-grained alignment, as reflected in more accurate symbol arrangements and more consistent reflective relationships. Taken together, these qualitative results suggest that explicitly promoting exploration beyond the model's default generation trajectory can improve the flexibility and accuracy of text–image alignment across both imaginative and realistic scenarios.

\subsection{Ablation Study}
\begin{table}
\centering
\small
\setlength{\tabcolsep}{4pt}
\caption{Ablation study of RA-GRPO. The table compares performance across different configurations: Baseline, with and without Counterfactual Synthesis (CF Synth.), and without Reflection. Higher values indicate better performance for all metrics.}

\begin{tabular}{p{1.5cm}cccc}
\toprule
Metric
& Baseline
& w/o CF Synth.
& w/o Reflection
& Ours \\
\midrule
HPSv2.1$\uparrow$        & 0.359 & 0.366 & 0.357 & \textbf{0.378} \\
PickScore$\uparrow$    & 0.227 & 0.226 & 0.229 & \textbf{0.230} \\
CLIPScore$\uparrow$   & 0.362 & 0.355 & 0.356 & \textbf{0.369} \\
ImageReward$\uparrow$ & 1.273 & 1.230 & 1.253 & \textbf{1.417} \\
HPSv3$\uparrow$       & 14.400& 15.245& 14.697& \textbf{15.324} \\
\bottomrule
\end{tabular}
\label{tab:ablation}

\end{table}

To assess the contribution of each core component, we conduct an ablation study on the FLUX backbone under the HPSv2.1 optimization setting. We examine two variants: w/o Counterfactual Synthesis, which removes the synthesized counterfactual transition and directly supervises the model using reflected samples, and w/o Reflection, which replaces the weak-to-strong difference vector with random Gaussian perturbations. The results in Table \ref{tab:ablation} show that both components are important. Removing Reflection leads to a clear drop in performance and brings the model close to the baseline, indicating that simple random perturbations are insufficient for effective exploration and that the weak-to-strong guidance plays a key role in steering optimization toward high-reward regions. Removing Counterfactual Synthesis slightly improves the target reward over the baseline, but degrades several generalization metrics, suggesting that directly optimizing discontinuous reflected states can introduce a mismatch between explored samples and the model’s native generation trajectory. Overall, the full RA-GRPO model achieves the best performance across all metrics, demonstrating that both directed reflection and counterfactual synthesis are necessary for robust reward optimization and generalized alignment.

\begin{figure*}[t!]
    \centering
    \includegraphics[width=0.9\textwidth]{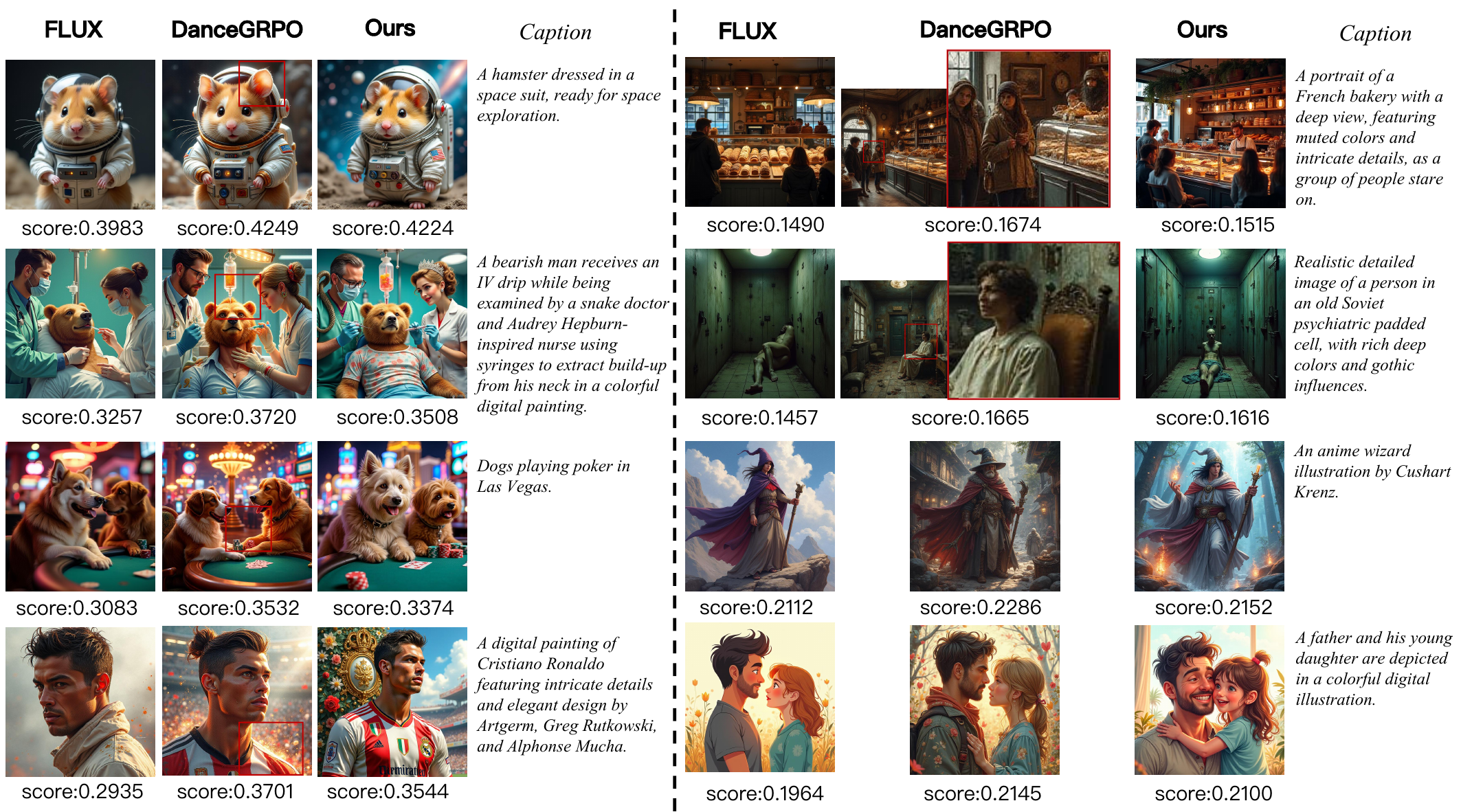}
    \caption{Qualitative comparison: The left panel shows results obtained with HPSv2.1-based optimization, while the right panel shows results obtained with PickScore-based optimization. Under HPSv2.1 training, DanceGRPO tends to introduce abnormal lighting and highlight artifacts, as illustrated in the red-boxed regions. Under PickScore training, DanceGRPO exhibits a tendency toward overly dark global tones; moreover, the zoomed-in red-boxed regions show that, despite dense high-frequency textures, object boundaries and structural contours remain blurry and poorly defined. }
    \label{fig:reward_hack}

\end{figure*}

\subsection{Human Preference Evaluation}
\textbf{\textit{Qualitative evidence of reward hacking.}}
 Fig.~\ref{fig:reward_hack} shows that DanceGRPO overfits reward-specific visual cues. When optimized with HPSv2.1, it introduces abnormal lighting, exaggerated highlights, and excessive local contrast; with PickScore, it favors darker tones and high-frequency textures with blurry boundaries and ambiguous structures. These artifacts increase reward-preferred saliency without improving semantic fidelity or perceptual realism. In contrast, Ours produces more natural lighting, clearer contours, and more coherent structures, despite slightly lower reward scores, indicating a better balance between reward optimization and perceptual quality. 

\textbf{\textit{User study.}}
We further compare DanceGRPO and Ours on 360 samples using randomized pairwise evaluation of text alignment, image quality, and aesthetic preference. Ours achieves win rates of 73.12\%, 69.35\%, and 67.91\%, respectively, demonstrating better alignment with human judgments despite possible discrepancies between proxy rewards and human preference.

\subsection{Optimal Configurations}
We study the reflection ratio \(r\), i.e., the fraction of sampled trajectories selected for Diffusion Reflection. As shown in Table~\ref{tab:reflection_ratio}, \(r=0.5\) achieves the best overall performance, whereas reflecting all trajectories (\(r=1.0\)) slightly degrades quality. Partial reflection promotes guided exploration while retaining unmodified trajectories for diverse and stable groupwise comparisons in GRPO. In contrast, reflecting every trajectory may reduce within-group contrast and weaken the relative learning signal. These results support \(r=0.5\) as the best trade-off between exploration and optimization stability.

\begin{table}
\centering
\caption{Effect of the reflection ratio \(r\) in Randomized Single-Step Reflection. Here, \(r\) denotes the fraction of sampled trajectories selected for reflection. }

\small
\setlength{\tabcolsep}{4pt}
\begin{tabular}{p{2.0cm}cccc}
\toprule
\multirow{2}{*}{\textbf{Metric} }
& \multicolumn{4}{c}{Reflection Ratio \(r\)} \\
\cmidrule(lr){2-5}
& 0.25 & 0.5 & 0.75 & 1.0 \\
\midrule
HPSv2.1 
& 0.369
& \textbf{0.378}
& \underline{0.377}
& 0.374
 \\

PickScore 
& 0.229
& \underline{0.230}
& \textbf{0.232}
& 0.230
 \\

CLIPScore 
& 0.363
& \textbf{0.369}
& \underline{0.367}
& 0.364
 \\

ImageReward 
& 1.392
& \textbf{1.417}
& \underline{1.406}
& 1.399
 \\

HPSv3 
& \underline{15.203}
& \textbf{15.324}
& 15.203
& 15.194 \\
\bottomrule
\end{tabular}

\label{tab:reflection_ratio}
\end{table}

\subsection{Extension to T2V Model}

\begin{table}[t]
\centering
\caption{Results on Wan2.1 under video-level and image-level reward optimization. Higher is better for all metrics.}
\label{tab:wan_image_video_results}
\resizebox{\linewidth}{!}{
\begin{tabular}{lccc}
\toprule
\textbf{Metric} & \textbf{Wan 2.1-1.3B} & \textbf{DanceGRPO} & \textbf{Ours} \\
\midrule
\multicolumn{4}{c}{\textbf{(A) Video-level setting (trained with VideoAlign-VQ)}} \\
\midrule
VQ \(\uparrow\) & 2.478 & 2.721 & \textbf{2.978} \\
MQ \(\uparrow\) & 0.177 & 0.239 & \textbf{0.303} \\
TA \(\uparrow\) & 0.153 & 0.235 & \textbf{0.297} \\
\midrule
\multicolumn{4}{c}{\textbf{(B) Image-level setting (first-frame only, trained with HPSv2)}} \\
\midrule
HPSv2.1 \(\uparrow\)    & 0.249 & 0.296 & \textbf{0.304} \\
PickScore \(\uparrow\)  & 0.210 & 0.216 & \textbf{0.219} \\
CLIPScore \(\uparrow\)  & 0.374 & 0.380 & \textbf{0.386} \\
ImageReward \(\uparrow\) & 0.471 & 0.817 & \textbf{0.932} \\
HPSv3 \(\uparrow\)      & 5.031 & 10.253 & \textbf{11.085} \\
\bottomrule
\end{tabular}
}

\end{table}

All experiments in Table~\ref{tab:wan_image_video_results} are conducted using the same data sources as the corresponding baselines to ensure a fair comparison. For the \textbf{video-level setting}, prompts are curated from the VidProM~\cite{wang2024vidprommillionscalerealpromptgallery} dataset. Wan2.1 is trained as a full video generator with VideoAlign-VQ as the reward function, and performance is evaluated using the VideoAlign metrics, including visual quality (VQ), motion quality (MQ), and text alignment (TA). Since video reward modeling is inherently more challenging due to the additional complexity of temporal dynamics and the relatively noisy supervision provided by learned video-level reward signals, we further introduce an \textbf{image-level setting} to verify that the effectiveness of our method does not depend on a particular video reward formulation or model architecture. Specifically, we use Wan2.1 as the backbone while treating it as an image generator, by considering only the first frame during both training and evaluation. In this setting, training prompts are drawn from the HPD dataset.

The results show that our method consistently outperforms both the vanilla Wan2.1-1.3B model and DanceGRPO under both evaluation protocols. In the video-level setting, our method achieves the best performance across all metrics. These improvements indicate that our method enhances not only static visual fidelity but also temporal coherence and semantic consistency in generated videos. In the image-level setting, our method again achieves the best results on all metrics, including HPSv2.1, PickScore, CLIPScore, ImageReward, and HPSv3. Notably, this setting removes the difficulty of video-level temporal reward modeling and evaluates only first-frame generation quality, thereby serving as a controlled test of whether our method generalizes beyond the original video reward setup. The consistent gains in both settings suggest that the proposed approach is not restricted to a specific reward model or task formulation; instead, it provides a more general and robust optimization benefit across both video and image generation regimes.

\section{Conclusion}

We introduced RA-GRPO, a framework that enhances reinforcement learning by integrating diffusion reflection mechanisms. By exploiting the discrepancy between weak and strong estimators to guide exploration and synthesizing counterfactual paths for training, RA-GRPO effectively rectifies sampling trajectories to navigate the optimization landscape more efficiently. This approach structurally addresses the cold-start dilemma by discovering high-probability modes in the data distribution early in training, while simultaneously mitigating reward hacking. Experiments across image and video generation demonstrate consistent improvements in visual quality, semantic alignment, and temporal coherence over existing reinforcement learning baselines. These findings establish reflection-driven exploration as a scalable paradigm for alignment, and future work may investigate diverse weak-to-strong estimator pairs to broaden the applicability of this framework.

\clearpage

\section*{Acknowledgment}

This work was supported in part by the Shenzhen Science and Technology Program
(Grant No.~RCJC20210706091946001), and in part by the Shenzhen Science and
Technology Program (Grant No.~ZDCY20250901104207008).

\bibliographystyle{ACM-Reference-Format}
\bibliography{sample-base}

\end{document}